\documentclass[runningheads]{llncs}
\usepackage{graphicx}
\usepackage{tikz}
\usepackage{comment}
\usepackage{amsmath,amssymb} 
\usepackage{color}

\usepackage{graphicx}
\usepackage{amsmath}
\usepackage{amssymb}
\usepackage{booktabs}

\usepackage{booktabs} 
\usepackage{tabularx}
\usepackage{multirow}
\usepackage{graphicx}
\usepackage{url}
\usepackage{array}
\usepackage{diagbox}
\usepackage{xcolor}
\usepackage{bbding}
\usepackage{caption}
\usepackage{lipsum}
\usepackage{wrapfig}

\usepackage{hyperref}
\usepackage{breakcites}

\usepackage[accsupp]{axessibility} 
\newcommand{\etal}{\textit{et al.~}}

\makeatletter
\g@addto@macro\@maketitle{
    \tiny
    \centering
    \includegraphics[width=1.\textwidth]{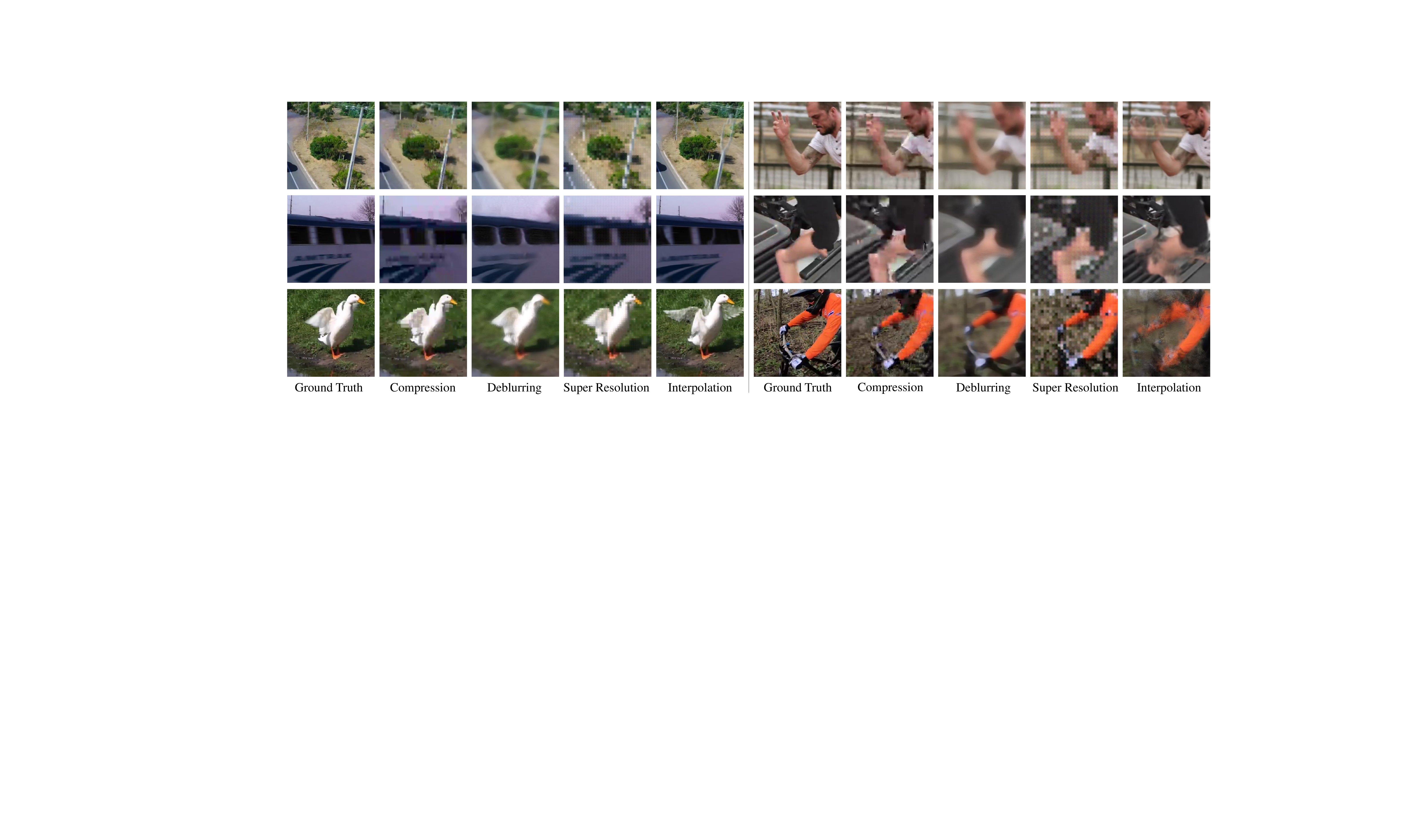}
   \captionof{figure}{Unique distortions in video frame interpolation results.
   }
\label{fig:fig1}
}
\makeatother

\begin{document}

\pagestyle{headings}
\mainmatter
\def\ECCVSubNumber{1542}

\title{A Perceptual Quality Metric for Video Frame Interpolation} 
\titlerunning{A Perceptual Quality Metric for Video Frame Interpolation}

\author{\vspace{-8pt}Qiqi Hou \and
Abhijay Ghildyal  \and
Feng Liu}
\authorrunning{Hou et al.}
\institute{Portland State University \\
\email{\{qiqi2,abhijay,fliu\}@pdx.edu}}

\maketitle

\begin{abstract}
Research on video frame interpolation has made significant progress in recent years. However, existing methods mostly use off-the-shelf metrics to measure the quality of interpolation results with the exception of a few methods that employ user studies, which is time-consuming. As video frame interpolation results often exhibit unique artifacts, existing quality metrics sometimes are not consistent with human perception when measuring the interpolation results. Some recent deep learning-based perceptual quality metrics are shown more consistent with human judgments, but their performance on videos is compromised since they do not consider temporal information. In this paper, we present a dedicated perceptual quality metric for measuring video frame interpolation results. Our method learns perceptual features directly from videos instead of individual frames. It compares pyramid features extracted from video frames and employs Swin Transformer blocks-based spatio-temporal modules to extract spatio-temporal information. To train our metric, we collected a new video frame interpolation quality assessment dataset. Our experiments show that our dedicated quality metric outperforms  state-of-the-art methods when measuring video frame interpolation results. Our code and model are made publicly available at \url{https://github.com/hqqxyy/VFIPS}.
\end{abstract}

\section{Introduction}
\label{sec:intro}
Video frame interpolation aims to generate frames between consecutive frames. It has attracted a lot of attention because of its wide applications in video editing~\cite{meyer2018deep}, video generation~\cite{kuroki2007psychophysical,kuroki2014effects}, optical flow estimation~\cite{long2016learning,wulff2018temporal}, and video compression~\cite{wu2018video}. Consequentially, great progress has been made and many video frame interpolation methods are  available~\cite{bao2019depth,choi2020channel,jiang2018super,kalluri2020flavr,lee2020adacof,li2020video,niklaus2018context,niklaus2020softmax,park2020bmbc,shi2022video,sim2021xvfi}.

To evaluate the quality of video frame interpolation results, most interpolation methods rely on traditional metrics, such as Peak Signal-to-Noise Ratio (PSNR) and Structural Similarity Index Measure (SSIM)~\cite{wang2004image}. These metrics estimate perceptual similarity between images by comparing the difference between pixels or carefully designed low-level visual patterns. While such methods achieve promising results, adapting them to measure video frame interpolation results is challenging. First, these metrics have been designed for general tasks such as video compression and other common low-level computer vision tasks, like super resolution, deblurring, and denoising, not specifically for video frame interpolation. Hence, these metrics are not optimal for measuring the quality of video frame interpolation results. As shown in Figure~\ref{fig:fig1}, video frame interpolation methods exhibit unique distortions. For example, the telegraph pole is twisted out of shape and the wings of the duck have ghosting artifacts. Second, traditional metrics rely on low-level features, such as per-pixel distance, and structure similarity~\cite{wang2004image}. However, human perception is highly complex, and hence, low-level features are insufficient for assessing the perceptual quality~\cite{zhang2018unreasonable}. These shortcomings motivate us to develop a dedicated quality metric for video frame interpolation.

Our research is inspired by the recent deep learning approaches to image quality assessment~\cite{bhardwaj2020unsupervised,czolbe2020loss,dendi2019full,ding2020iqa,kim2017deep,kim2018deep,korhonen2019two,korhonen2020blind,li2019quality,talebi2018nima,xu2020c3dvqa,zhang2018unreasonable}. Deep learning approaches, such as LPIPS~\cite{zhang2018unreasonable}, take an image and its reference image as input and output the perceptual score. Instead of comparing low-level features, they learn perceptual similarity from the internal features of deep convolutional networks. Such methods have shown great success for image perceptual quality assessment. However,  applying these metrics to video frame interpolation results frame by frame ignores the temporal information in the videos, which often make these metrics inconsistent with human perception.

This paper presents a learned perceptual video similarity metric dedicated to video frame interpolation results. It takes an interpolated video and its reference ground-truth video as input and outputs the perceptual similarity between them. It builds upon state-of-the-art image perceptual quality metrics but extends to videos, leading to a naturally spatio-temporal architecture. Specifically, our network extracts the features from each frame. Then, we design a dedicated spatio-temporal module to capture the spatio-temporal information from videos. This spatio-temporal module consists of the Swin Transformer blocks~\cite{liu2021swin}, which can explore the interaction among different regions, learn to attend the important regions, and extract spatio-temporal video information. 

To train our network, we collected a large Video Frame Interpolation Perceptual Similarity (VFIPS) dataset that contains 25,887 video triplets. Each video triplet contains two interpolated videos, their reference video, and the corresponding perceptual preference. Our dataset consists of various artifacts generated by eleven state-of-the-art video frame interpolation methods~\cite{bao2019depth,bao2019memc,choi2020channel,jiang2018super,lee2020adacof,li2020video,liu2019deep,niklaus2018context,niklaus2020softmax,niklaus2017video,park2020bmbc}. With this dataset, we train our video perceptual metric using Binary Cross Entropy (BCE) as the loss function in an end-to-end manner. Our experiments show that our metric outperforms existing metrics significantly when measuring video frame interpolation results.

This paper contributes to the research on video frame interpolation and perceptual quality assessment by 1) providing the first video perceptual similarity metric dedicated to video frame interpolation, 2) designing a novel neural network architecture for video perceptual quality assessment based on the Swin Transformers, and 3) building a large video frame interpolation perceptual similarity dataset.

\section{Related Work}
\label{sec:related} 
This paper investigates quality assessment of video frame interpolation results. Below, we first briefly survey video frame interpolation methods. Then we discuss traditional video quality metrics and recent deep-learning-based approaches. We also discuss video quality assessment datasets and vision transformers. 

\textbf{Video frame interpolation.} Video frame interpolation aims to estimate  intermediate frames between two consecutive frames. They can be classified into three categories: kernel-based methods~\cite{niklaus2017video,niklaus2017videoSep,niklaus2020revisiting,niklaus2021revisiting,xu2019learning}, phase-based methods~\cite{meyer2018phasenet,meyer2015phase}, and flow-based methods~\cite{bao2019depth,bao2019memc,cheng2021multiple,choi2021motion,danier2022st,hu2022many,huang2020rife,jiang2018super,lee2020adacof,li2020video,liu2019deep,liu2017video,niklaus2018context,niklaus2020softmax,park2020bmbc,park2021asymmetric,raket2012motion,reda2019unsupervised,sim2021xvfi,xue2019video,yu2021training}. Kernel-based methods~\cite{niklaus2017video,niklaus2017videoSep,niklaus2020revisiting,xu2019learning} estimate a kernel for each pixel in the frame indicating the weights for its neighboring pixels to synthesize the new pixel. They generate the synthesized frame in a local convolutional manner. Phase-based methods~\cite{meyer2018phasenet,meyer2015phase} generate intermediate frames by per-pixel phase modification. Currently, most of methods are flow based~\cite{bao2019depth,bao2019memc,jiang2018super,lee2020adacof,li2020video,liu2019deep,liu2017video,niklaus2018context,niklaus2020softmax,park2020bmbc,raket2012motion,reda2019unsupervised,sim2021xvfi,xue2019video}. Typically, their networks contain two modules: optical flow module and frame synthesized module. The optical flow module is used to estimate the optical flow between input frames. By warping the input frame based on the estimated optical flow, the frame synthesized module synthesizes the intermediate frames. Most of the video frame interpolation methods use PSNR and SSIM as their evaluation metrics. Some methods~\cite{lee2020adacof,niklaus2020softmax,xu2019quadratic} also explore Interpolation Error (IE)~\cite{baker2011database}, Normalized Interpolation Error (NIE)~\cite{baker2011database}, and Learned Perceptual Image Patch Similarity (LPIPS)~\cite{zhang2018unreasonable} to measure their results. However, these metrics are insufficient to evaluate the perceptual similarity. Traditional quality metrics, such as PSNR, only use low-level features and LPIPS lacks temporal information. In this paper, we propose a learned video quality metric dedicated to video frame interpolation.

\textbf{Video quality metrics.} Based on the availability of the reference video, traditional video quality metrics can be classified into No-Reference (NR), Reduced-Reference (RR), and Full Reference (FR). Due to the limitation of space, we refer readers to a comprehensive survey for traditional methods~\cite{winkler2008evolution}. Our method is most related to the FR methods, which have access to the reference video. Traditional FR methods infer the perceptual similarity from the differences from pixels or low-level features~\cite{van1996color,gardiner1997development,hamada1999picture,lubin1997human,VMAF,watson1998toward,watson2001digital,webster1993objective,winkler2005digital,wolf1997objective,wu2017digital,xiao2000dct}. For instance, Wange~\etal observed that the human visual system (HVS) is adaptive to structural similarity~\cite{wang2004image}. They evaluated the perceptual similarity from the structural deformity between a reference and distortion image. In comparison to SSIM, Pinson ~\etal had a better correlation with HVS on video quality assessment~\cite{pinson2004new}. They also considered multiple perceptual-based features, including blurring, unnatural motion, noise, color distortion, and block distortion.  Li~\etal proposed a multi-method metric to fuse multiple video perceptual quality metrics using a support vector machine~\cite{li2016toward}. It can preserve the strength of the individual metrics and achieve promising results. However, these metrics only use low-level features, which is insufficient for many nuances of human perception. Recently, with the success of deep learning, many research works introduce the deep learning method to video quality assessment~\cite{chen2020learning,tu2021ugc}. However, these metrics are designed for the traditional artifacts like video compression, while our method focuses on artifacts that arise from video frame interpolation.

\textbf{Deep learning based quality metrics.} Our method is most related to the recent deep learning based metrics for image quality assessment~\cite{bhardwaj2020unsupervised,czolbe2020loss,dendi2019full,ding2020iqa,kim2017deep,kim2018deep,korhonen2019two,korhonen2020blind,li2019quality,talebi2018nima,xu2020c3dvqa,zhang2018unreasonable}. Instead of using low-level features, these methods learn features from deep convolutional networks and train a network in an end-to-end manner. For full-reference image quality assessment (FR-IQA), Kim~\etal estimated the perceptual similarity using a deep network~\cite{kim2017deep}. Their method takes a distorted images and an error map as input and estimates the sensitivity map. They get the subjective score by multiplying the sensitivity map with the error map. Zhang~\etal explored the convolutional features from a classification network~\cite{zhang2018unreasonable}. By comparing the features extracted using the convolutional network, their method is able to predict the perceptual similarity robustly. They also built a large-scale dataset for image perceptual quality assessment, which contains traditional distortions and CNN-based distortions. Their dataset also contains distortions from frame interpolation. However, it is designed for image quality assessment and only contains single images. In contrast, our dataset contains videos for frame interpolation, which enable us to learn the temporal information. Ding~\etal proposed an image quality metric with tolerance to texture resampling~\cite{ding2020iqa}. Bhardwaj~\etal trained their metric using the information theory-guided loss function~\cite{bhardwaj2020unsupervised}. Czolbe~\etal trained an image quality metric for GAN using Watson’s perceptual model~\cite{czolbe2020loss}. While these methods have achieved great success on image quality assessment, directly adapting these methods to videos will be disadvantageous to these methods because they ignore the spatio-temporal consistency. In contrast, our work builds upon these deep learning methods and considers the spatio-temporal information in our spatio-temporal network module, thus providing a dedicated solution to video frame interpolation quality assessment.

\textbf{Video quality datasets.} Traditional video quality datasets focus on video compression and transmission~\cite{antkowiak2000final,bampis2018towards,ghadiyaram2017live,ghadiyaram2017subjective,min2020study,VMAF,seshadrinathan2010study}. Video Quality Experts Group (VQEG) dataset was proposed for the secondary distribution of television~\cite{antkowiak2000final}. Most videos in the VQEG dataset are interlaced. To address this problem, Kalpana~\etal built the Laboratory for Image and Video Engineering (LIVE) Video Quality Database~\cite{bampis2018towards,seshadrinathan2010study}. It is designed for the H.264 compression, MPEG-2 compression, and video transmission. In their recent works, they extended the dataset for the distortion in the streaming~\cite{ghadiyaram2017live,min2020study}. Li~\etal proposed a video quality dataset for video compressions, including H.264/AVC, HEVC, and VP9~\cite{VMAF}. Different from these datasets, our work focuses on distortions from video frame interpolation. Recently, Danier~\etal proposed the BVI-VFI dataset for video frame interpolation~\cite{danier2022subjective}. It is a relatively small dataset and hence not suitable for training deep learning models. Therefore, we make use of it as a test dataset to compare our method against various traditional and recently proposed metrics.

\textbf{Vision Transformer.} There has been considerable interest in vision transformers due to their impressive performance in image classification~\cite{dosovitskiy2020image,li2021localvit,liu2021swin}, object detection~\cite{carion2020end}, and image restoration~\cite{chen2021pre}. Recently, the vision transformers have also been introduced for the image quality assessment \cite{cheon2021perceptual,you2021transformer}. Their networks combined vision transformer~\cite{dosovitskiy2020image} and CNN, and achieved promising results for image quality assessment. Our work adapts the recent proposed Swin Transformer~\cite{liu2021swin} for video quality assessment.

\section{Video Frame Interpolation Quality Dataset}
\label{sec:data}
We collected a Video Frame Interpolation Perceptual Similarity (VFIPS) dataset. It consists of a wide variety of  distortions from video frame interpolation. Each sample consists of two videos synthesized from different interpolation methods, its reference video, and its perceptual judgments $h\in\{0, 0.33, 0.66, 1\}$. We can denote it as $\{\mathbf{V}_A, \mathbf{V}_B, \mathbf{V}_R, h\}$. Below, we briefly describe how we prepare and annotate this dataset.

\subsection{Data Preparation}
\label{sec:source}

\textbf{Source videos}. From YouTube, we collected 96 videos under the Creative Commons Attribution license (reuse allowed), which enables us to share this dataset with the community. Among them, 45 videos are 120 fps, and 51 videos are 60 fps. Most videos last over 10 minutes. The resolutions of the source videos are either 1080p or 4K. We downsampled these videos to 540p with the bicubic downsampling function from OpenCV. From these videos, we extracted video clips for quality assessment. After eliminating the video clips containing cut shots or interlaced frames, we curate, in total, 23,856 12-frame video clips. These 12-frame sequences serve as a reference in our dataset. 

\textbf{Videos synthesized by frame interpolation methods}. We generated the synthesized videos for each reference video with 11 state-of-the-art video frame interpolation methods, including SepConv~\cite{niklaus2017video}, Super-SloMo~\cite{jiang2018super}, CtxSyn~\cite{niklaus2018context}, CyclicGen~\cite{liu2019deep}, MEMC-Net~\cite{bao2019memc}, DAIN~\cite{bao2019depth}, CAIN~\cite{choi2020channel}, RRIN~\cite{li2020video}, AdaCoF~\cite{lee2020adacof}, BMBC~\cite{park2020bmbc}, and SoftSplat~\cite{niklaus2020softmax}. We set $\times 4$ scale for 120-fps videos and $\times 2$ scale for 60-fps videos. 

For each reference sequence, we need to compare two video frame interpolation results. To make it easy to view and find differences between videos, we keep the size of the cropped video patches as $256\times256$. Specifically, we calculated the mean $\ell_1$ error map between two synthesized videos and selected the patch location with the highest error by sliding patch windows.

\subsection{Annotation}
\label{sec::anno}

Compared to annotating images, annotating videos often requires much greater effort from users. For instance, a video contains many frames, but the distortions might only show in one or just a few frames. In such cases, a user needs to play the video multiple times to analyze differences and make a final judgment. In our case, a user often takes more than 1 minute to annotate 1 sample. Hence, can we make use of some existing perceptual quality metrics to help us annotate more samples?

Recent developments in perceptual quality metrics show that considerable progress has been made with regard to correlating these metrics with human judgment~\cite{bhardwaj2020unsupervised,czolbe2020loss,ding2020iqa,kim2017deep,talebi2018nima,zhang2018unreasonable}. The result of a single image quality assessment can provide helpful prior knowledge for the quality assessment of videos. For instance, by visual inspection, we found that if all frames of one video have significantly better quality than the frames in another video, this video often has a better quality. Based on this observation, we make use of the widely adopted state-of-the-art image quality metric LPIPS~\cite{zhang2018unreasonable} to help annotate samples where two videos have significantly different LPIPS scores. These LPIPS annotated examples are only used for training and not used for testing. While they are not perfect,  they can be used to train a good VFI quality metric. A similar strategy was also adopted in the PIEAPP metric~\cite{Prashnani_2018_CVPR}. For samples which is hard to make a judgment via LPIPS, we collect human annotations.

\noindent\textbf{Automatic annotation.} We select 500 video triplets for our study to test the hypothesis that \emph{when the difference in LPIPS scores of two videos is significant, the judgment via LPIPS is consistent with the human judgment}. We collect human annotations for these 500 triplets. Each video triplet contains one reference video and two synthesized videos. For each synthesized video, we calculate its LPIPS score as the mean of the LPIPS scores of the frames in that video. We tested whether the judgments via LPIPS are consistent with the human judgments. Evidence from this study suggests that the judgment via LPIPS can be reliably used for annotating triplets with large LPIPS difference.

\begin{figure*}[t]
    \scriptsize
    \centering
        \includegraphics[width=1.0\textwidth]{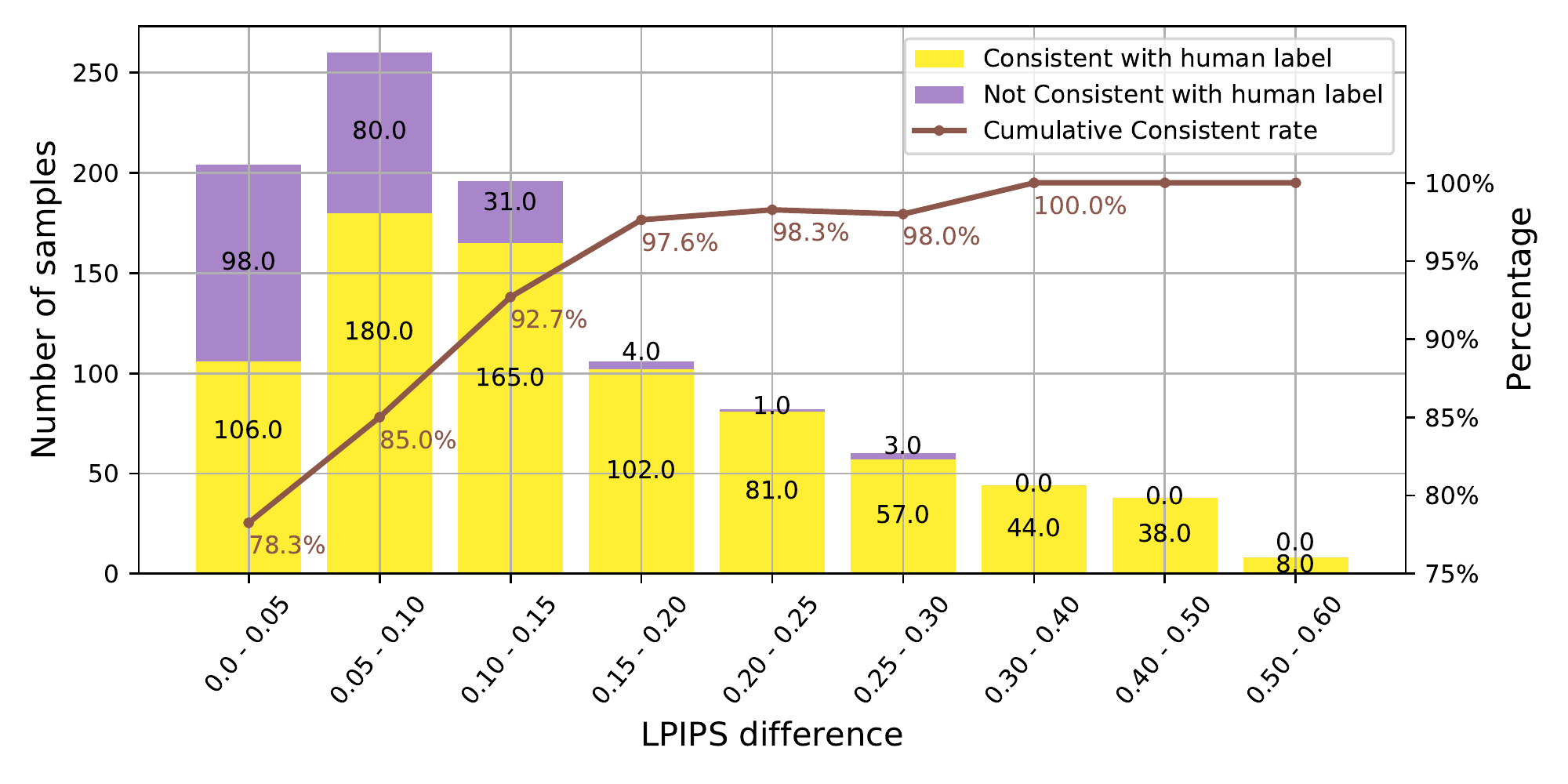}
    \caption{LPIPS difference vs. human labeling result. }
    \label{fig:lpips_diff}
    \vspace{-0.1in}
\end{figure*}

Figure~\ref{fig:lpips_diff} shows the LPIPS difference versus the human judgment, where the LPIPS difference is the absolute difference between the LPIPS scores of a pair of videos. The judgment via LPIPS is consistent with the human judgment with the increase in the LPIPS difference. We can find that 97.6\% judgments via LPIPS are consistent with the human judgments when the LPIPS difference is greater than 0.15. Based on this observation, we select 0.15 as the LPIPS-difference threshold. If a new video triplet's LPIPS difference is greater than 0.15, we take the LPIPS judgment as its ground truth label. For the rest of the video triplets where LPIPS difference is less than 0.15, we collect human annotations. It helps save time and effort involved in the recruitment and collection of high-quality human annotations. In the end, for our dataset, we have 18,939 video triplets with ground truth labels annotated via LPIPS. 

\noindent\textbf{Manual annotation.} For the samples where LPIPS-difference is less than 0.15, we collect judgments on perceptual quality by humans.

\begin{figure*}[t]
    \scriptsize
    \centering
        \includegraphics[width=0.85\textwidth]{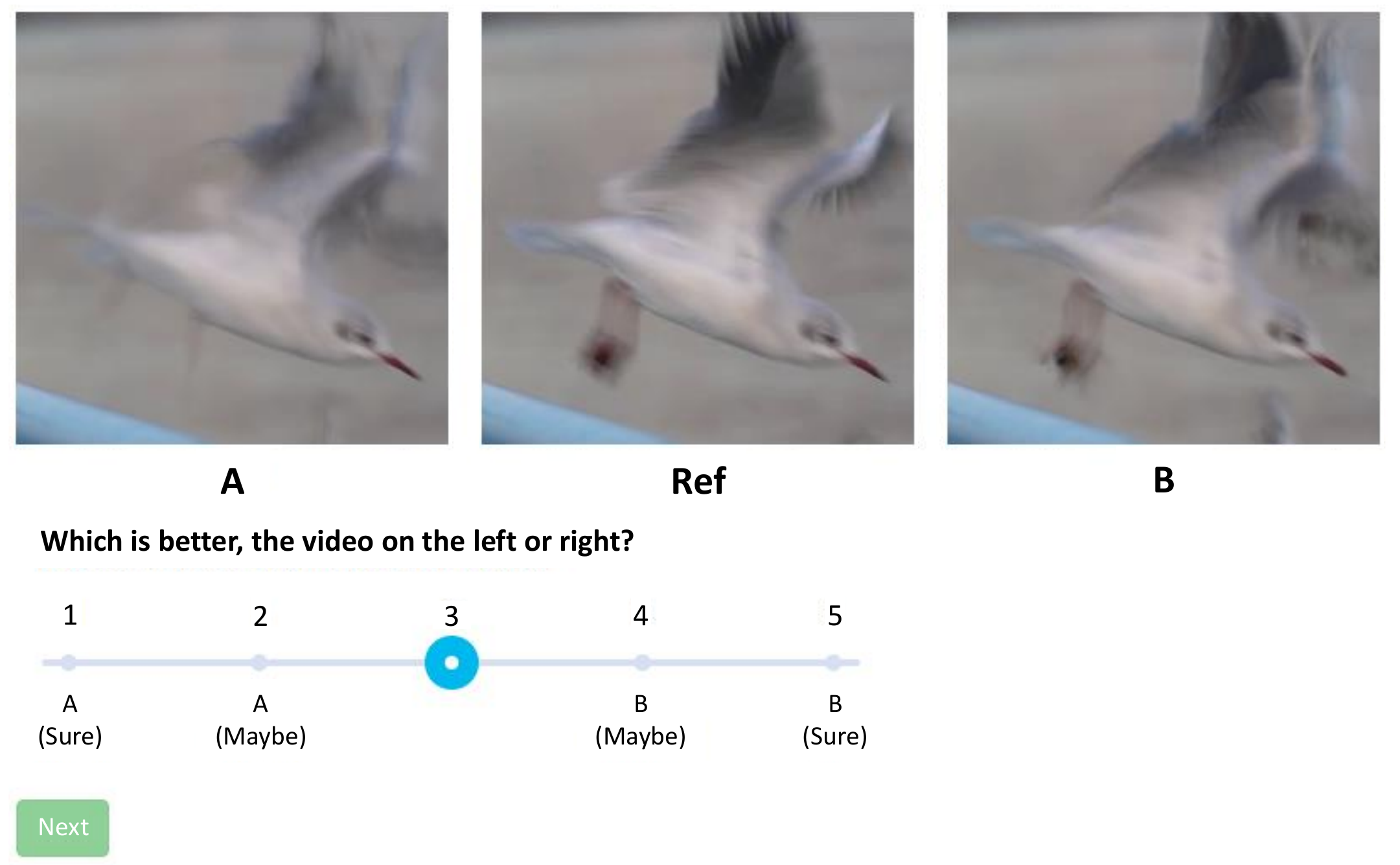}
    \caption{User interface. Two frame interpolation results and one reference are played simultaneously in a loop.}
    \label{fig:interface}
    \vspace{-0.1in}
\end{figure*}

\textbf{User interface.} For the annotation, we presented participants with a reference video and two corresponding video frame interpolation results. The videos were played simultaneously to help participants compare the two synthesized videos. For judgment, participants were asked which of the two synthesized videos has a higher perceptual quality. We played the videos in a loop without interruption or human intervention. Figure~\ref{fig:interface} shows the user interface. It has four options \{``A(Sure)'', ``A(Maybe)'', ``B(Maybe)'', ``B(Sure)"\}. Please note that the middle point in Figure~\ref{fig:interface} is not an option.

\textbf{Playback setting.} To find the best playback setting, we tested video clips of length from 4 to 16 frames. If a video is too short, human annotators would not get enough temporal information to make a decision. But a too-long video often leads to ambiguity. For example, the quality of a video can be high in the first half but be poor in the last half. In such cases, it is difficult for users to make a decision. We empirically found that a video clip with 12 frames works well for users to annotate the quality. We also explored the playback speed in our study. At a high fps, such as 8 fps, users will find it extremely hard to compare the distortions, as each frame only shows for a very short time. However, a low fps will make it harder to judge the temporal consistency. We empirically found that 2 fps works well.

We collected annotations on 5,948 video triplet samples from 41 participants. Following LPIPS~\cite{zhang2018unreasonable}, each sample was labeled by 3 participants, and the mean of the judgments was taken to decide which video had a higher quality. All participants that we recruited were volunteers from diverse backgrounds. For our training dataset, we randomly select 5,353 samples out of the 5,948 human-annotated video triplets and all 18,939 sample triplets with judgment via LPIPS. The rest 595 human-annotated samples were used for validation.

\section{Video Frame Interpolation Quality Metric}
\label{sec:method}
Our method takes a video $\mathbf{V} = \{\mathbf{I}_0, \mathbf{I}_1, \cdots, \mathbf{I}_N \}$ and its reference video $\mathbf{V}_R = \{\mathbf{I}_{R, 0}, \mathbf{I}_{R, 1}, \cdots, \mathbf{I}_{R, N} \}$ as inputs and estimates the perceptual similarity $d$ between these two videos, where $N$ is the number of frames. 

\begin{figure*}[t]
    \scriptsize
    \centering
        \includegraphics[width=1.\textwidth]{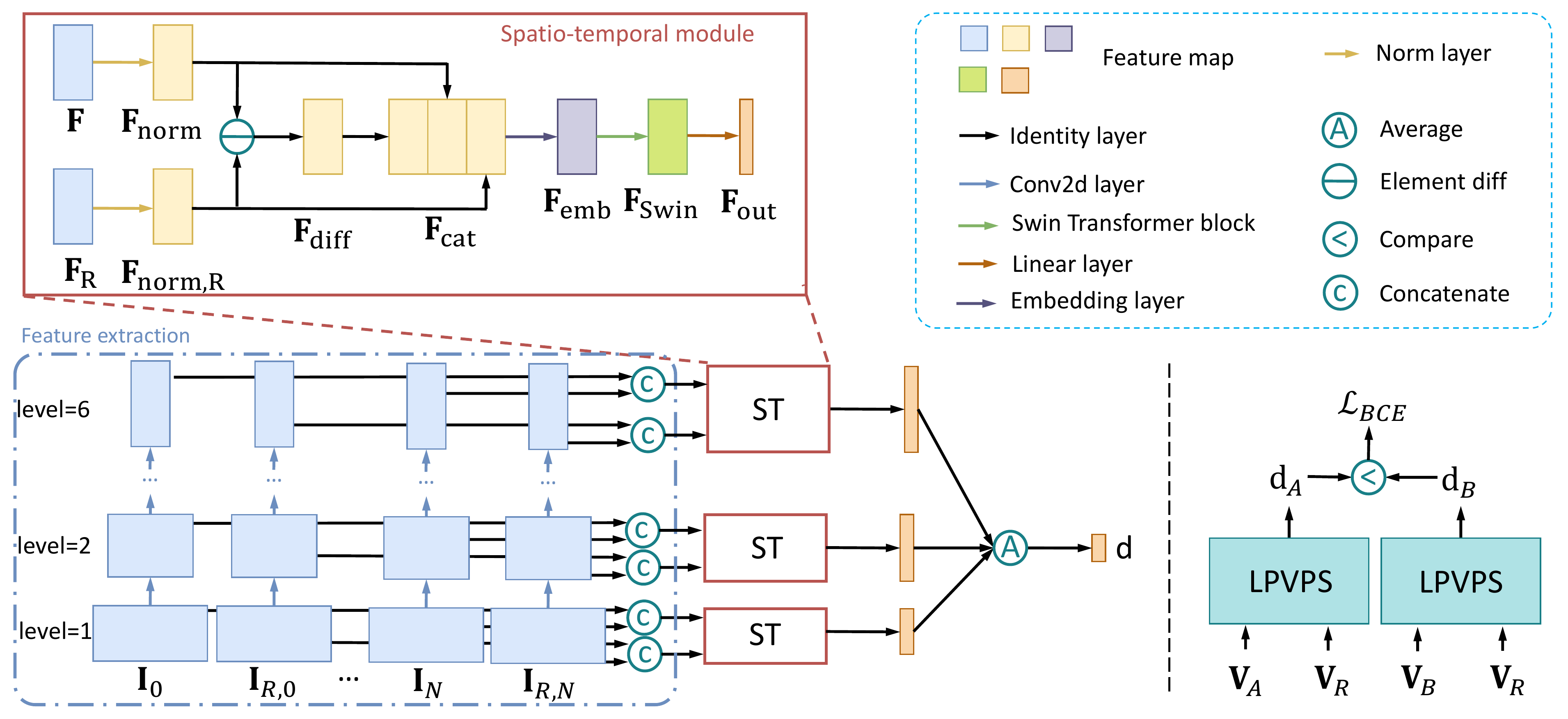}
    \caption{Our network architecture. (Left) Learned Perceptual Video Patch Similarity (LPVPS) takes a video $\mathbf{V}$ and its reference $\mathbf{V}_R$ as input and predicts their perceptual similarity $d$. (Right) Our network is trained in a Siamese manner.}
\label{fig:arch}
\vspace{-0.1in}
\end{figure*}

Figure~\ref{fig:arch} shows the architecture of our network. Our network first extracts the feature maps for each image in the videos. For each feature map, we design a spatio-temporal (ST) module to capture the spatio-temporal information and estimate the perceptual similarity of features. Finally, we predict the perceptual similarity between the input video $\mathbf{V}$ and its reference $\mathbf{V}_R$ by averaging the similarity across all features. Our network is trained in a Siamese manner. Below we describe our network in more detail.

\textbf{Feature extraction.} As shown in Figure~\ref{fig:arch}, we extract features from the video $\mathbf{V}$ and its reference video $\mathbf{V}_R$. Specifically, we build a pyramid network with five levels. Each level contains two $3\times3$ \texttt{Conv2D} layers. The stride of the second \texttt{Conv2D} layer is set to 2 to have a large valid receptive field. From the first to the fifth level, the numbers of channels are 16, 32, 64, 96, and 128. It can be represented as follows.
\begin{equation}
    \{\mathbf{F}_{level, i}\}_{level=1}^L = f_{ext}(\mathbf{I}_i),
\end{equation}
where $f_{ext}(\cdot)$ indicates the operation of the extraction network. $\mathbf{F}$ indicates the extracted features. $L$ indicates the number of feature levels. The feature extraction network shares weights for all frames and videos. For each level, we concatenate the features from all frames as the final features,
\begin{equation}
    \mathbf{F}_{level} = f_{cat}(\mathbf{F}_{level, 0}, \cdots, \mathbf{F}_{level, N}),
\end{equation}
where $f_{cat}(\cdot)$ is the concatenation operation. $N$ is the number of frames.

\textbf{Spatio-temporal (ST) module} is designed to measure the perceptual distance between the feature maps $\mathbf{F}$ and its reference $\mathbf{F}_R$. It first estimates the difference between two normalized features as follows.
\begin{equation}
     \mathbf{F}_{diff} = \lVert f_{norm}(\mathbf{F}) - f_{norm}(\mathbf{F}_R)\rVert_1,
\end{equation}
where $f_{norm}(\cdot)$ indicates the operation of unit normalization in the channel dimension~\cite{zhang2018unreasonable}. As reported in previous work on image perceptual metrics~\cite{goodman1972seven,markman2005nonintentional,medin1993respects}, the perceptual similarity is not only related to the difference between two images, but also the content of images. Therefore, our network leverages both the feature difference $\mathbf{F}_{diff}$ and the source features.
\begin{equation}
     \mathbf{F}_{cat} = f_{cat}(\mathbf{F}_{diff}, f_{norm}(\mathbf{F}), f_{norm}(\mathbf{F}_R)).
\end{equation}

As shown in Figure~\ref{fig:arch}, we first adopt a linear embedding layer to project the features $\mathbf{F}_{cat}$ to a fixed dimension 32.
\begin{equation}
     \mathbf{F}_{emb} = f_{emb}(\mathbf{F}_{cat}),
\end{equation}
where $f_{emb}(\cdot)$ indicates the linear embedding layer that consists of a \texttt{Conv2D} layer with a $1\times1$ kernel.

We adopt a Swin Transformer block~\cite{liu2021swin} to learn the temporal information across frames. Compared to the original design of Swin Transformer block~\cite{liu2021swin}, we do not use \texttt{Layer Norm (LN)} layers, which is critical to measure the difference between two videos. \texttt{LN} layers normalize the differences between two features and compromise the accuracy of the network. We empirically find that the dimension of the embeded features as 32, the head number as 2, and the window sizes of 4 can get good results and require relatively less computation and memory.

ST modules are applied to the features at different levels, which can capture the spatio-temporal information at multiple scales. It is similar to the pyramid architectures of optical flow networks~\cite{liu2019selflow,sun2018pwc}. The final distance is calculated by averaging the outputs at all the $L$ levels from ST modules.
\begin{equation}
     d = \frac{1}{LN}\sum_{l=0}^L \sum_{n=0}^N \mathbf{F}_{Swin, l}^n
\end{equation}
where $N$ indicates the number of elements in $\mathbf{F}_{Swin, l}$.

\textbf{Loss function.} As shown in Figure~\ref{fig:arch}, our network is trained in a Siamese manner. Given two videos with different distortions $\mathbf{V}_A$,  $\mathbf{V}_B$, its reference video $\mathbf{V}_R$, and its judgement $\hat{h}$, we first predict the perceptual similarity $d_A$ and $d_B$ from ($\mathbf{V}_A$, $\mathbf{V}_R$) and ($\mathbf{V}_B$, $\mathbf{V}_R$), respectively. We calculate the probability $p$ with a sigmoid layer,
\begin{equation}
    p = sigmoid(d_A - d_B).
\end{equation}
Then, we calculate the binary cross entropy loss $\mathcal{L}_{BCE}$ as:
\begin{equation}
    \mathcal{L}_{BCE} = -(\hat{h} \log(p) + (1 - \hat{h}) \log (1 - p)).
\end{equation}

\textbf{Training.} We use PyTorch to train our neural network. Following \cite{zhang2018unreasonable}, our learning rate is set to 0.0001. We use a mini-batch size of 8 and train the network for 20 epochs. We randomly resize the videos in the scale of [0.5, 1]. Our network is randomly initialized and uses AdamW~\cite{kingma2014adam} as the optimizer. 

\section{Experiments}
\label{sec:exp}
\begin{table}[t]
    \centering
    \caption{Comparison with state-of-the-art methods.}
    \label{table:comp_others}
    \small
    \begin{tabular}{clcccc}
            \toprule
            \multicolumn{2}{c}{\multirow{1}[2]{*}{Method}} & VFIPS (val.) & \multicolumn{3}{c}{BVI-VFI~\cite{danier2022subjective} (test)}\\ \cline{4-6}
            \multicolumn{2}{c}{} &2AFC & SROCC & PLCC &KROCC\\ \midrule
            \multirow{7}[2]{*}{Image} 
            & PSNR &0.763 & 0.742 & 0.722 &0.656\\
            & SSIM~\cite{wang2004image} & 0.784&0.739 & 0.746 &0.639\\
            & MS-SSIM~\cite{wang2003multiscale} &0.794 & 0.772 & 0.789 &0.667\\
            & LPIPS (VGG)~\cite{zhang2018unreasonable} &0.808 & 0.628  & 0.796 &0.517\\
            & DISTS~\cite{ding2020iqa} &0.801 &0.597  &0.763  &0.517\\
            & PIM-1~\cite{bhardwaj2020unsupervised} &0.787 &0.492  &0.668 &0.428 \\ 
            & Watson-DFT~\cite{czolbe2020loss} &0.800 & 0.628 &0.706 &0.538\\ \midrule
            \multirow{3}[0]{*}{Video} 
            & STRRED~\cite{bovik2017robust} &0.777 &0.614 &0.682 &0.539  \\ 
            & VMAF~\cite{VMAF} &0.805 &0.583 &0.614 &0.483\\
            & DeepVQA~\cite{tencentDVQA} &0.588 &0.369 &0.271 &0.300 \\
            & VSFA~\cite{li2019quality} &0.660 &0.108 &0.486 &0.050\\
            &Ours & \textbf{0.830} & \textbf{0.794}  & \textbf{0.870} &\textbf{0.700}\\ 
            \bottomrule
    \end{tabular}
    \vspace{-0.1in}
\end{table}

We use the BVI-VFI~\cite{danier2022subjective} dataset as our test set. It contains 36 reference videos at 3 different frame rates: 30fps, 60fps, and 120fps.  All the videos last 5 seconds. Each reference video has 5 distorted videos generated from different video frame interpolation algorithms, including frame repeating, frame averaging, DVF~\cite{liu2017video}, QVI~\cite{xu2019quadratic} and ST-MFNet~\cite{danier2022st}. Please note that video frame interpolation methods used in the BVI-VFI dataset are different from the ones used in our VFIPS dataset. To test our metric on the BVI-VFI dataset, we evaluate our metric in a sliding-window manner and take the average score. We selected Spearman Rank Order Correlation Coefficient (SROCC), Pearson Linear Correlation Coefficient (PLCC), and Kendall Rank Order Correlation Coefficient (KROCC) scores as our evaluation metrics. We calculated the scores for each reference video and reported the mean score as the final result. We also report our results on our VFIPS validation set. Following LPIPS~\cite{zhang2018unreasonable}, we report the Two Alternative Forced Choice (2AFC) scores~\cite{zhang2018unreasonable}. We compare our method to the representative state-of-the-art metrics and conduct ablation studies to evaluate our metric. 

\subsection{Comparisons to Existing Metrics}

\begin{table}[t]
\setlength{\tabcolsep}{4.8pt}
        \scriptsize
        \centering
        \caption{Comparison on the X-TEST(4K) dataset~\cite{sim2021xvfi}.}
        \label{table:comp_xtest}
        \begin{tabular}{cccccccc}
            \toprule
            Method & PSNR & SSIM~\cite{wang2004image} & MSSSIM~\cite{wang2003multiscale} &LPIPS~\cite{zhang2018unreasonable} &STRRED~\cite{bovik2017robust} &VMAF~\cite{VMAF} & Ours \\ \midrule
            2AFC &0.752 &0.637 &0.737 &0.748 &0.722 &0.735 &\textbf{0.789} \\
            \bottomrule
        \end{tabular}
        \vspace{-0.1in}
\end{table}

We compare our method to both the state-of-the-art image perceptual metrics, including PSNR, SSIM~\cite{wang2004image}, MS-SSIM~\cite{wang2003multiscale}, LPIPS(VGG)~\cite{zhang2018unreasonable}, DISTS~\cite{ding2020iqa}, PIM-1~\cite{bhardwaj2020unsupervised}, and Watson-DFT~\cite{czolbe2020loss}, and the state-of-the-art video perceptual metrics, including VMAF~\cite{VMAF}, DeepVQA~\cite{kim2017deep}, and STRRED~\cite{soundararajan2012video}. We also compare with the Non-Reference metrics, VSFA~\cite{li2019quality}.  For each method, we obtained the results by using the official codes/models provided by their authors, except for DeepVQA which we adopt its re-implementation from Tencent~\cite{tencentDVQA}. For the image perceptual similarity metrics, we evaluate the score per frame and take the average as the final score for that video sequence.

\begin{figure*}[t]
    \scriptsize
    \centering
        \includegraphics[width=0.98\textwidth]{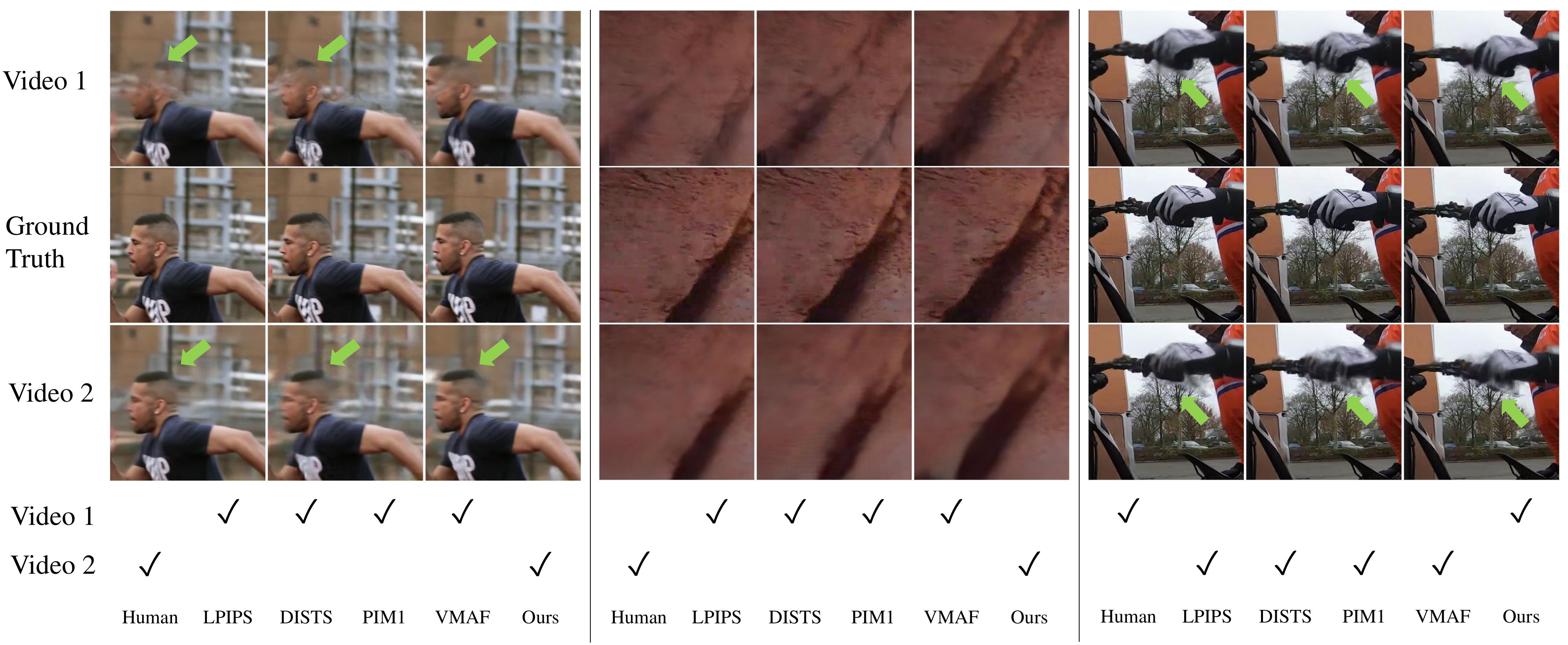} 
    \caption{Visual examples on the VFIPS dataset. Green arrows are used to label the area with noticeable difference. We mark the preference of each method using ``\Checkmark''. Compared to other methods, our method is consistent with humans.}
\label{fig:comp_others}
\vspace{-0.1in}
\end{figure*}

\begin{figure*}[t]
    \scriptsize
    \centering
        \includegraphics[width=0.98\textwidth]{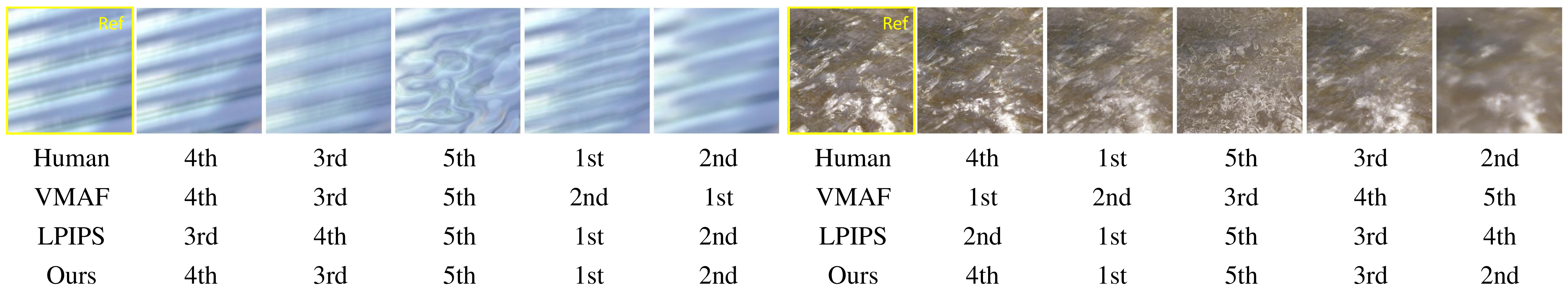}
    \caption{Visual examples on the BVI-VFI dataset~\cite{danier2022subjective}. Yellow rectangles are used to show the reference video. We report the rank for the distorted videos. Compared to other metrics, our metric is more consistent with humans.}
\label{fig:comp_bvivfi}
\vspace{-0.2in}
\end{figure*}

As shown in Table~\ref{table:comp_others}, our method outperforms the state-of-the-art methods by a considerable margin on both the VFIPS dataset and the BVI-VFI dataset. Especially when compared to the image-based similarity methods such as LPIPS, our method provides significant gains in terms of 2AFC on the VFIPS dataset and SROCC, PLCC, and KROCC on the BVI-VFI dataset. We attribute this improvement to the spatial-temporal module that captures spatio-temporal information. Compared to the video-based methods, our method again achieves significant performance gains on all metrics on both datasets, specifically, 0.025 in terms of 2AFC on the VFIPS validation set and 0.180 in SROCC, 0.188 in PLCC, 0.161 in KROCC on the BVI-VFI dataset. Figure~\ref{fig:comp_others} shows the visual examples on the VFIPS validation set. The predictions of our method are more consistent with humans. In the first example, Video 2 suffers less distortions to the person's head. In the second example, Video 2 is temporally consistent. In the third example, the hand in Video 2 is distorted more. As indicated, our predictions are consistent with humans in these examples. Figure~\ref{fig:comp_bvivfi} shows the visual examples from the BVI-VFI dataset~\cite{danier2022subjective}. For each reference video, it shows the rank for the distorted videos. Our predictions are more consistent with humans.

We also conduct our experiments on the X-TEST dataset~\cite{sim2021xvfi}, which has 15 testing videos with 4K resolution. We use SepConv, Super-Slomo, AdaCof, and XVFI to generate VFI results. Following Sim~\etal\cite{sim2021xvfi}, we set the temporal distance of 32 frames for the large motion. For each video, we interpolate 32 frames and take the middle 12 frames for testing. There are 90 pairs in total. Each pair is annotated by 3 users. As reported in Table~\ref{table:comp_xtest}, our method outperforms other methods on such a large-motion dataset.

\subsection{Ablation Studies}
\label{sec:ablation}

\begin{table}[t]
\setlength{\tabcolsep}{5pt}
    \centering
    \caption{Effectiveness of different feature extraction network on the BVI-VFI datasets~\cite{danier2022subjective}. }
    \label{table:comp_arch}
    \begin{tabular}{cccccc}
            \toprule
            Extractor & SROCC & PLCC & KROCC &Param(M) &Runtime(ms) \\ \midrule
            AlexNet~\cite{krizhevsky2012imagenet} &0.761 &0.832 &0.650 &14.5 &12.8 \\
            I3D~\cite{carreira2017quo} &0.659 &0.758 &0.550 &20.3 &33.2\\
            Ours-3D &0.728  &0.738 &0.639 &8.6 &13.6 \\ \midrule
            Ours-2D & \textbf{0.794} &\textbf{0.870} &\textbf{0.700} &\textbf{4.6} & \textbf{10.4}\\
            \bottomrule
    \end{tabular}
\end{table}

\textbf{Feature extraction network.} We investigate the impact of architectures of the feature extraction network in Table~\ref{table:comp_arch}. Following LPIPS~\cite{zhang2018unreasonable}, we replace our extraction network with several classic architectures, including AlexNet~\cite{krizhevsky2012imagenet}, I3D~\cite{carreira2017quo},  and ours-3D network. We use the official implementation of AlexNet and I3D~\cite{carreira2017quo}. For ours-3D, we replace the \texttt{Conv2D} layers in our feature extraction module with \texttt{Conv3D} layers with $3\times3\times3$ kernels. We measure the runtime for a 12-frame $256\times256$ video on a single Nvidia RTX A5000 GPU.

As shown in Table~\ref{table:comp_arch}, our method not only achieves the best performance but also needs the least parameters. Compared to AlexNet, our feature extraction network achieves better performance while our method only needs 34.1\% parameters and is $1.23\times$ faster. Compared to I3D and ours-3D, our network achieves much better results and has less parameters. We attribute the improvement to that our feature extraction network keeps more temporal information. Specifically, I3D and ours-3D downsample the features temporally, while our network keeps features of all the frames. The temporal compression in the feature extraction network might hurt the performance of the whole metric.

\begingroup
\setlength{\intextsep}{0.05in}
\setlength{\columnsep}{0.2in}

\begin{wraptable}{r}{0.6\textwidth}
        \setlength{\tabcolsep}{4.8pt}
        \centering
        \caption{Effectiveness of the spatio-temporal module on the BVI-VFI dataset~\cite{danier2022subjective}.}
        \label{table:comp_st}        
        \begin{tabular}{cccc}
            \toprule
            ST Module & SROCC & PLCC & KROCC \\ \midrule
            None &0.617 &0.663 &0.539 \\
            \texttt{Conv3D}  &0.761 &0.819 &0.661 \\ 
            Original Swin &0.728 &0.766 &0.639 \\ 
            Ours-Swin w. \texttt{LN} &0.724 &0.746 &0.611 \\ \midrule
            Ours-Swin & \textbf{0.794} &\textbf{0.870} &\textbf{0.700}\\
            \bottomrule
        \end{tabular}
\end{wraptable}

\textbf{Spatio-temporal module.} We study the spatio-temporal module in Table~\ref{table:comp_st}. We use different blocks for the spatio-temporal module. ``\texttt{Conv3D}'' indicates the \texttt{Conv3D} layers with $12\times3\times3$ kernels. ``Original Swin'' indicates the official Swin Transformer blocks. ``Ours-Swin w. \texttt{LN}'' indicates our Swin Transformer block with \texttt{LN} layers. Compared to the original Swin, it contains fewer parameters, including 32 vs. 96 (channels), 4 vs. 7 (window size). It has slightly worse performance. As shown in Table~\ref{table:comp_st}, original Swin and ours-Swin w. \texttt{LN} do not perform as well as Conv3D. We attribute this performance loss to the \texttt{LN} normalizing the differences between features. Removing the \texttt{LN} layers improves the results by 0.070 on SROCC, 0.124 on PLCC, and 0.089 on KROCC on the BVI-VFI dataset. 

\endgroup

\begingroup
\setlength{\intextsep}{0.in}
\setlength{\columnsep}{0.2in}

\begin{wraptable}{r}{0.6\textwidth}
        \setlength{\tabcolsep}{7.5pt}
        \centering
        \caption{Effectiveness of annotations on the BVI-VFI dataset~\cite{danier2022subjective}}
        \label{table:comp_anno}
        \begin{tabular}{cccc}
            \toprule
            Annotations & SROCC & PLCC & KROCC \\ \midrule
            Human &0.719 &0.753 &0.611 \\
            Automatic  &0.653 &0.687 &0.567 \\ 
            All & \textbf{0.794} &\textbf{0.870} &\textbf{0.700}\\
            \bottomrule
        \end{tabular}
\end{wraptable}

\textbf{Automatic annotations by LPIPS.} We study the impact of annotations in Table~\ref{table:comp_anno}. As discussed in Section~\ref{sec::anno}, our training set contains two parts: samples obtained through automatic annotation or human labeling. We train our network with different training subsets: human annotations, automatic annotations, and a combination of both. As shown in Table~\ref{table:comp_anno}, the combined set achieves the best performance. The automatic training data can increase the diversity of the training set and improve the generalization capability of the metric. 

\endgroup

\section{Limitations and Future Work}
\label{sec:lim}
Many image-based quality assessment metrics can generate an error map between a distorted image and its reference to analyze altered pixels. However, our metric cannot produce such an error map for individual frames in the video since our spatio-temporal module contains convolutional layers that fuse features from these frames at an early stage in the inference. In the future, it will be interesting to design a network that can generate such error maps for each frame to provide more information on the distortion within and across the frames.

Recently, image perceptual quality metrics, such as LPIPS~\cite{zhang2018unreasonable}, are used as loss to create visually pleasant images or videos~\cite{niklaus2021learned}. In theory, our network can also be used as a loss function to train a video frame interpolation network to produce visually pleasant videos. However, our metric has a high requirement for GPU memory. It takes about 2500 Mb memory to process one $256 \times 256$ 12-frame image sequence, which might become a bottleneck in training a video frame interpolation network. In the future, we aim to reduce the network size and study how our video frame interpolation quality metric will help optimize video frame interpolation networks as a learned perceptual loss.

\section{Conclusion}
This paper presented a video perceptual quality metric for video frame interpolation. Our metric first extracts pyramid features for individual frames in the videos. Then it compares features at each level using a spatio-temporal module to capture the spatio-temporal information. The spatio-temporal module is composed of the Swin Transformer blocks to capture the spatio-temporal information. We also collected a dataset for video frame interpolation. Our annotations were from humans and a widely adopted image perceptual quality metric. Our experiments showed that our dedicated video quality metric outperforms existing metrics for assessing video frame interpolation results.

\noindent\textbf{Acknowledgment} This work was made possible in part thanks to Research Computing at Portland State University and its HPC resources acquired through NSF Grants 2019216 and 1624776. Source frames in Figure~\ref{fig:fig1}, \ref{fig:interface}, and \ref{fig:comp_others}  are used under a Creative Commons license from Youtube users Ignacio, Scott, Animiles, H-Edits, 3 Playing Brothers, TristanBotteram, popconet, and billisa.

\clearpage

\bibliographystyle{splncs04}
\bibliography{egbib}
\end{document}